\documentclass[12pt]{article}
\usepackage{amssymb}
\usepackage[ruled,vlined]{algorithm2e}
\usepackage{natbib,graphicx,setspace,lscape,longtable}
\usepackage{natbib,epsfig,graphicx}
\usepackage{booktabs}
\usepackage{rotating}
\usepackage{multirow}
\usepackage{bm}
\usepackage{amsmath,amsthm,amssymb,color}
\usepackage{wrapfig} 
\usepackage{array}
\newcolumntype{P}[1]{>{\centering\arraybackslash}p{#1}}
\newcolumntype{M}[1]{>{\centering\arraybackslash}m{#1}}
\newcolumntype{L}{>{\centering\arraybackslash}m{2cm}}
\usepackage{floatrow}
\newfloatcommand{capbtabbox}{table}[][\FBwidth]
 \usepackage{array, diagbox}

\bibpunct{(}{)}{;}{a}{,}{,}
\usepackage{hyperref}
\allowdisplaybreaks
\hypersetup{
     colorlinks   = true,
  linkcolor=blue,
            urlcolor=blue,
            citecolor=blue
}

\usepackage{xr}

\newtheorem{theorem}{Theorem}
\newtheorem{lemma}{Lemma}

\def\be{\begin{equation}}
\def\ee{\end{equation}}
\def\ben{\begin{equation*}}
\def\een{\end{equation*}}
\def\bea{\begin{eqnarray}}
\def\eea{\end{eqnarray}}
\def\bd{\begin{displaymath}}
\def\ed{\end{displaymath}}
\def\bda{\begin{eqnarray*}}
\def\eda{\end{eqnarray*}}

\def\lsk{\left(}
\def\rsk{\right)}

\def\beq{\begin{equation}}
\def\eeq{\end{equation}}
\def\beqr{\begin{eqnarray}}
\def\eeqr{\end{eqnarray}}
\def\beqrs{\begin{eqnarray*}}
\def\eeqrs{\end{eqnarray*}}
\def\bet{\begin{theorem}}
\def\eet{\end{theorem}}
\def\bel{\begin{lemma}}
\def\eel{\end{lemma}}

\def\bg{\begin{figure}[tbph]\begin{center}}
\def\eg{\end{center}\end{figure}}

\def\bc{\begin{center}}
\def\ec{\end{center}}

\def\1{\mbox{\boldmath $1$}}

\def\SD{\mbox{SD}}

\def\lsk{\left(}
\def\rsk{\right)}

\numberwithin{equation}{section}

\begin{document}

\begin{center}
{\bf\Large Unaware Fairness: Hierarchical Random Forest for Protected Classes}\\
\bigskip
{Xian Li }\\
{\textit{ The Australian National University}}

{\it }
 
\author{%
  Xian Li u5708983 \\
}

\end{center}

\begin{abstract}

Procedural fairness has been a public concern, which leads to controversy when making decisions  with respect to protected classes, such as race, social status and disability. 
Some protected classes can be inferred according to some safe proxies like
surname and geolocation for race. Hence, implicitly utilizing the predicted protected classes based on the related proxies when making decisions is an efficient approach to  circumvent this issue and seek just decisions. In this article, we propose a hierarchical random forest model for prediction without explicitly involving protected classes. 
 Simulation experiments are conducted to show
the performance of hierarchical random forest model. An example is analyzed from Boston police  interview records to illustrate the usefulness of the proposed model.

\end{abstract}

\section{Introduction}

The United States is one of the most ethnic diverse countries in the world. Thus any unfair decisions regarding  protected classes including race and ethnicity can result into huge controversial debate. For example, \cite{larson2016we} questioned the racial fairness of  the algorithm: Correctional Offender Management Profiling for Alternative Sanctions (COMPAS), which is a case management and decision support tool used by U.S. courts to assess the likelihood of a defendant becoming a recidivist. \cite{larson2016we} found that blacks are almost twice as likely as whites to be labeled a higher risk but not actually re-offend.

On the other hand, the U.S. Commission on Civil Rights banned  police stations to "redlining" suspects
based on certain racial groups, e.g., unreasonable traffic stop on civilians due to
factors which show the discrimination to certain civil groups. 
Therefore, any statistical prediction based on racial profile would also be controversial. 
Consequently, many surveys or records have eliminated the collection of racial information. 
However, characteristics such as race, age, and socio-economic class determine other features about us that are relevant to the outcome of some performance tasks. These are protected attributes, but they are still critical to certain tasks.

To resolve this issue, a variety of methods have been proposed. For instance, \cite{chen2019fairness} employed safe proxies to impute
these missing protected variables for statistical prediction. Motivated by this idea, this article proposes a hierarchical random forest to make uncontroversial prediction without explicitly involving protected classes. In the bottom layer of random forest, we adopt proxy variables to predict the protected classes and in the top layer we make the ultimate prediction based on the predicted protected classes along with other factors. We also provide the prediction interval follow the spirit of \cite{meinshausen2006quantile}. Results are found that hierarchical random forest  perform even better in some aspects compared to the naive random forest using protected classes directly.

The rest of the paper is organized as follows. In section \ref{Meth}, we introduce the model and prediction interval.  Simulations are conducted to validate our model in Section \ref{NumStud}. Section \ref{real} elucidate the usefulness of our model by analyzing Boston police interview data. Concluding remarks are present in Section \ref{Conc}

\section{Methodology}\label{Meth}

\subsection{Model}
Random forests are non-parametric and often  accurate and robust  to explore a variety of data. Moreover, random forest has been propose for missing data imputation (e.g., see \citealt{stekhoven2015missforest}). Thus, we chose random forest to capture the relation between protected classes and other covariates. Our proposed model can be written as
\[\hat Y^{(1)}=\hat{f}^{(1)}(\cdot, \hat Y^{(2)}),\]
\[\hat Y^{(2)} = \hat{f}^{(2)}(\cdot).\]
where $\hat{f}^{(1)}(\cdot)$ and  $\hat{f}^{(2)}(\cdot)$ refers to the top and bottom layer  random forest methods and $\cdot$  and $\hat Y^{(2)}$ refers to the covariates. In our case $\hat Y^{(1)}$ refers to the desired predicted result and $\cdot$ refers to the covariates except protected classes such as location and date etc. $\hat Y^{(2)}$ refers to the predicted protected classes. In this model, we treat protected classes as latent. As a result,  protected classes like race does not explicitly show as covariates.
For random forest's  the number of predictors $m$, we consider  the number of drawn candidate variables in each split equals to $\sqrt{p}$. Also,  we consider the case where the drawn observations are with  replacement.Moreover, the node size parameter specifies the minimum number of observations in a terminal node. Setting it lower leads to
trees with a larger depth which means that more splits are performed until the terminal nodes. Since we have a large sample size, we would set a larger node size to decrease the computational burden. We employ the algorithm proposed by \cite{breiman2001random} to optimise the model in this article.

\subsection{Prediction Interval}

The uncertainty of the random forest predictions can be estimated using several approaches including U-statistics approach of \cite{mentch2016quantifying} and monte carlo simulations approach of \cite{coulston2016approximating}. 
Prediction intervals in this paper are constructed by Quantile Regression Forests (\citealt{meinshausen2006quantile}). The prediction of
random forests can then be seen as an adaptive neighborhood classification and regression
procedure (\citealt{lin2006random}). For every covariate, a set of weights for the
original  observations is obtained. It is illustrated that 
that random forest not only a good approximation to the conditional
mean but to the full conditional distribution. The prediction of random forests is equivalent to the weighted mean of the observed response variables. For
quantile regression forests, trees are grown as in the standard random forests algorithm. The
conditional distribution is then estimated by the weighted distribution of observed response
variables.  We only have a brief introduction to Quantile Regression Forests in this article; see \cite{meinshausen2006quantile} for a more thorough treatment.

\section{Simulation Studies}\label{NumStud}

\subsection{Simulation Design}

To evaluate the performance of the proposed model, we conduct Monte Carlo simulation studies with various settings. We follow \cite{zhu2015reinforcement}'s work to create three simulation scenarios that represent different
aspects, which usually arise in machine learning. Such aspects
include, classification problem 
and nonlinear structure. For each scenario, we further consider generate  independent test samples to calculate the performance measure. Each simulation is repeated $B=100$
times, and the averaged performance is presented. The
simulation settings are as follows: We simulated $N$ observations of Normal distribution. Specifically,the
simulation settings are as follows:
$x_1$ and $x_2$ are independently generated from uniform distribution $\mathcal{U}\lsk0,1\rsk$, where $x_3$ is computed regarding to $0.4*x_1+0.4*x_2+0.2*\mathcal{U}\lsk0,1\rsk$.

\textit{Scenario 1: Linear.}
$Y = 3x_1+3x_2+2x_3+\varepsilon$, where $(\cdot)^+$ represents the positive part

\textit{Scenario 2: Non-Linear.}
$Y = 100(x_1 -0.5)^2(X_2-0.25)^++cos(x_3)+\varepsilon$, where $(\cdot)^+$ represents the positive part.

\textit{Scenario 3: Classification.}
Let $\mu=\Phi\lsk10\times (x_1-1)+10\times|x_2-0.5|+10*x_3\rsk,$ where $\Phi$ denotes a normal c.d.f.
Draw $Y$ independently from  Bernoulli($\mu$).

 The above model settings satisfy our technical conditions. For each of the above settings, a total of $N=500$ samples are used. 
\subsection{Performance Measurement}\label{measure}
For a convincing evaluation, we take several measures for comparison. 
We present the bias $\frac{\sum_{i=1}^N\hat Y_i-Y_i}{N}$, standard deviation $\frac{\sqrt{\sum_{i=1}^N\lsk\hat Y_i-\frac{\sum_{i=1}^N\hat Y_i}{N}\rsk^2}}{N}$ and mean square error (MSE) $\frac{\sum_{i=1}^N(\hat Y_i-Y_i)^2}{N}$ for numerical results and confusion matrix for categorical results respectively. For each replication, we obtain $\textrm{bia}_b,\textrm{\SD}_b,\textrm{MSE}_b$ for $b=1,\ldots,B$ and report the average of them. 
\subsection{Simulation Results}
\begin{figure}[H]
\begin{floatrow}
\ffigbox{%
\includegraphics[scale=0.27]{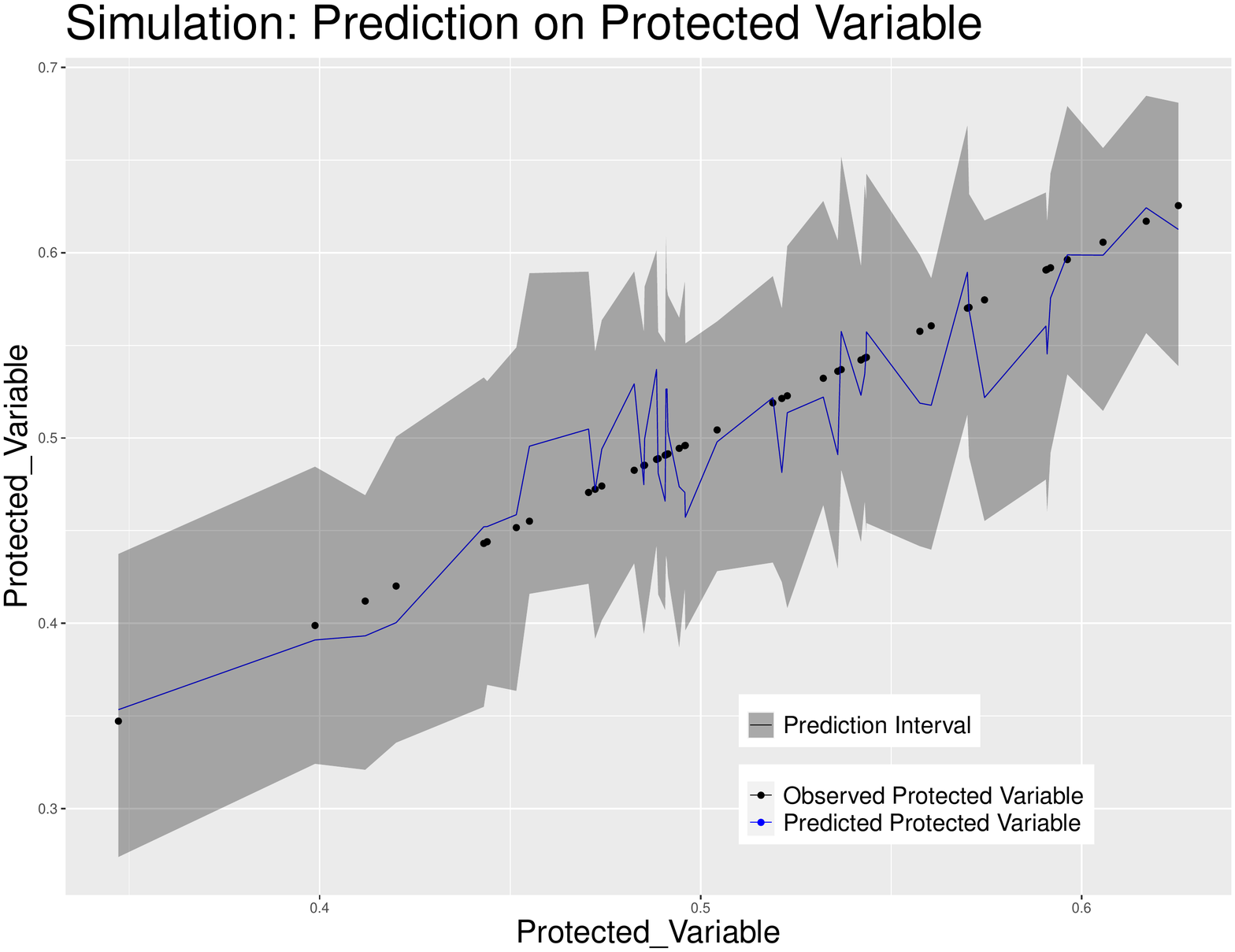}
}{%
  \caption{Prediction of Protected Class}
  \label{protect:fig}
}
\ffigbox{%

\includegraphics[scale=0.27]{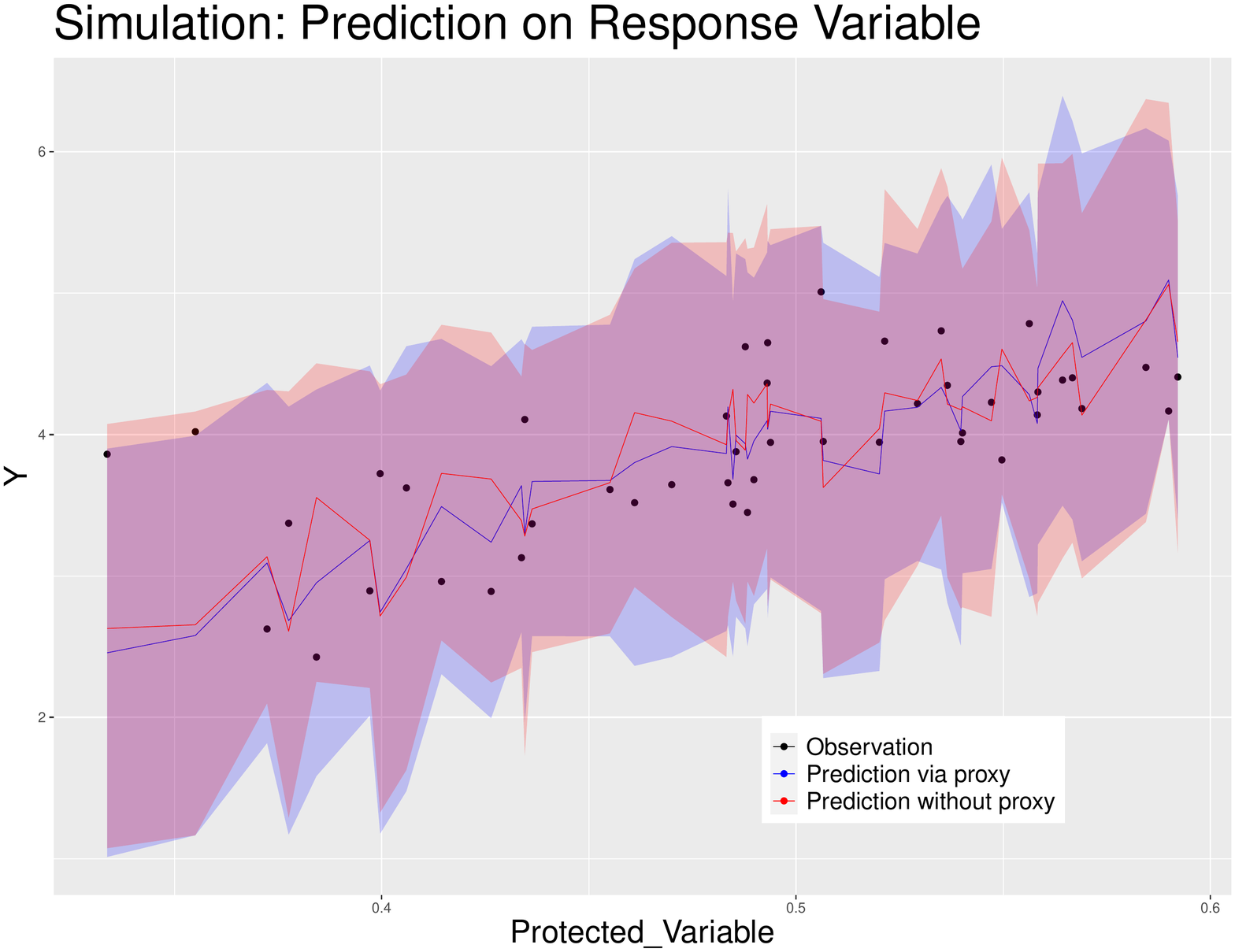}

}{%
  \caption{Linear} 
  \label{linear:fig}
}
\end{floatrow}
\end{figure}

\begin{figure}[H]
\begin{floatrow}
\ffigbox{%
\includegraphics[scale=0.27]{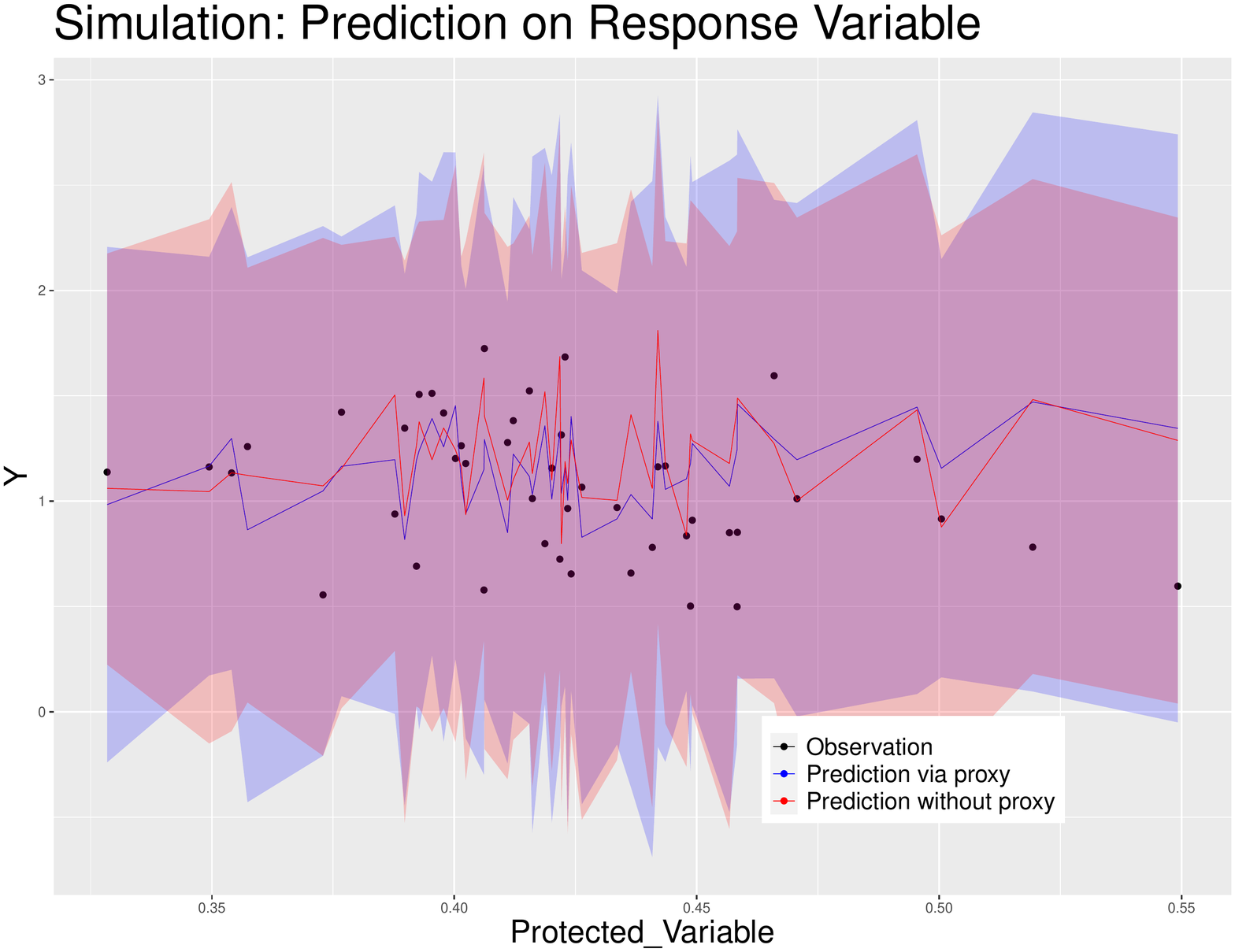}
}{%
  \caption{Non-Linear}
  \label{nonlinear:fig}%
}
\ffigbox{%

\includegraphics[scale=0.27]{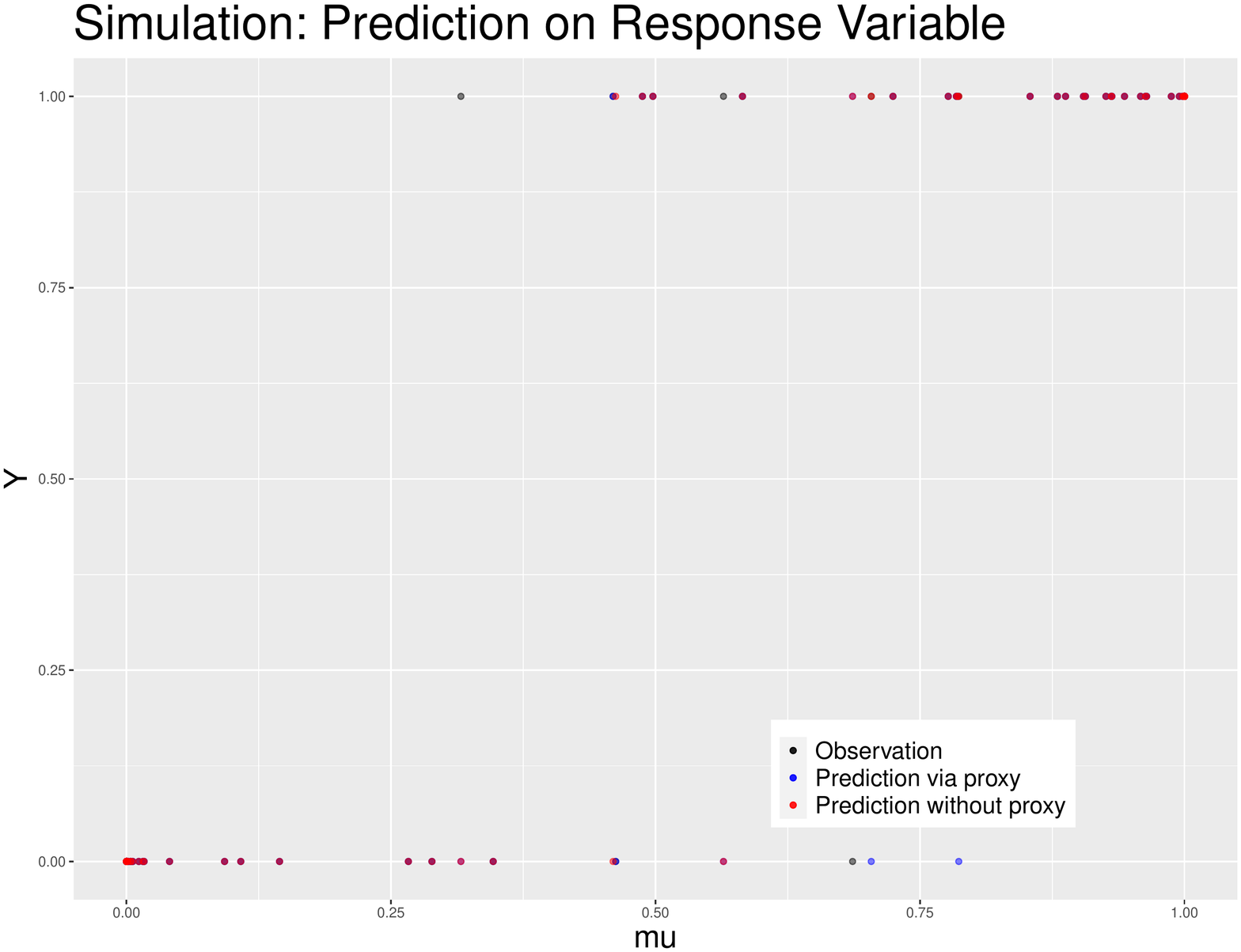}

}{%
  \caption{Classification}
  \label{class:fig}
}
\end{floatrow}
\end{figure}

Table \ref{linear:tab}-\ref{wtih:tab} summarize testing sample prediction error for each simulation setting. We compare the proposed model (with proxy) with naive random forest (without proxy) using protected classes directly. There is clear evidence that
the hierarchical  random forest produce a reasonable prediction under
these settings. For linear settings, the proposed random forest even outperforms naive random forest in terms of bias and standard deviation. This could be caused by the fact that hierarchical random forest correct its predecessor as boosting. The corresponding visualized predicted pattern are shown in Figure \ref{linear:fig}-\ref{class:fig}. The black dots are the observed data, lines are predictions and  ribbons are the corresponding 90\% prediction interval. As we can see that the prediction standard deviation of both models for Non-Linear settings are  larger. Overall, two models' performance are comparable.

\begin{figure}[!htbp]
\begin{floatrow}
\capbtabbox{%
\small
\begin{tabular}{c|c|cc} 
\hline\hline 
&without proxy&with proxy &\\ [0.5ex]
\hline
bias &-0.0043 &0.0353&\\ 
\SD& 1.5504 &1.5263\\
MSE &5.8402 &5.7654\\ 
[1ex] 
\hline\hline 
\end{tabular}
}{%
  \caption{Linear} \label{linear:tab}
}
\capbtabbox{%
\small
\begin{tabular}{c|c|cc} 
\hline\hline 
&without proxy&with proxy &\\ [0.5ex]
\hline
bias &0.0052 &0.0064&\\ 
\SD&0.4732& 0.5031 &\\
MSE &1.3412& 1.3749&\\ 
[1ex] 
\hline\hline 
\end{tabular}

}{%
  \caption{Non-Linear} \label{nonlinear:tab}
}
\end{floatrow}
\end{figure}

\begin{figure}[!htbp]
\begin{floatrow}
\capbtabbox{%
\small
\begin{tabular}{c|c|cc} 
\hline\hline 
&Actual P.&Actual N. &\\ [0.5ex]
\hline
Predicted P. &  94.63\%& 9.64\%&\\ 
Predicted N.&5.36\%& 90.35\%&\\
[1ex] 
\hline\hline 
\end{tabular}

}{%
  \caption{Confusion matrix without proxy}
  \label{wtihout:tab}
}
\capbtabbox{%
\small
\begin{tabular}{c|c|cc} 
\hline\hline 
&Actual P.&Actual N. &\\ [0.5ex]
\hline
Predicted P. & 93.77\%& 10.13\%&\\ 
Predicted N.&6.22\%&89.86\% &\\
[1ex] 
\hline\hline 
\end{tabular}

}{%
  \caption{Confusion matrix with proxy} 
\label{wtih:tab}
}
\end{floatrow}
\end{figure}

\section{Real Data Analysis}\label{real}
\subsection{Problem Setup}\label{setup}

We first present an overview of our dataset. 
Police field interview data in Boston include 34 variables such as Location, Subject Race, Sex, Reason, Clothing, Vehicle and Officer Race etc. Those data were collected from January 2001 to January 2020 and stored in a 155230 $\times$  34 data matrix. In our analysis, we use the 10 major variables which includes \textbf{Sex},
\textbf{Location: Street},
     \textbf{Location: District}, 
  \textbf{Location: City},
     \textbf{Incident Date}, 
     \textbf{Priors}, 
     \textbf{Race}, 
     \textbf{Skin Complexion}, 
    \textbf{Clothing}, 
     \textbf{Incident Reason }. Among these covariates,
\textbf{Race} is considered as the protected class in our analysis. As a result, our main objectives are to (1)  predict racial profile with a set of factors by bottom-level random forest, (2)  predict \textbf{Incident Reason} with predicted latent \textbf{Race}  and (3) predict the daily occurrence of police field interview with predicted latent \textbf{Race} Therefore, the racial profile would not be an explicit covariate. We also will compare the prediction accuracy of the hierarchical model with the naive  one and discuss the fairness of both models.
\subsection{Preliminary Analysis}
\begin{figure}[H]
\begin{floatrow}
\ffigbox{%
\includegraphics[scale=0.25]{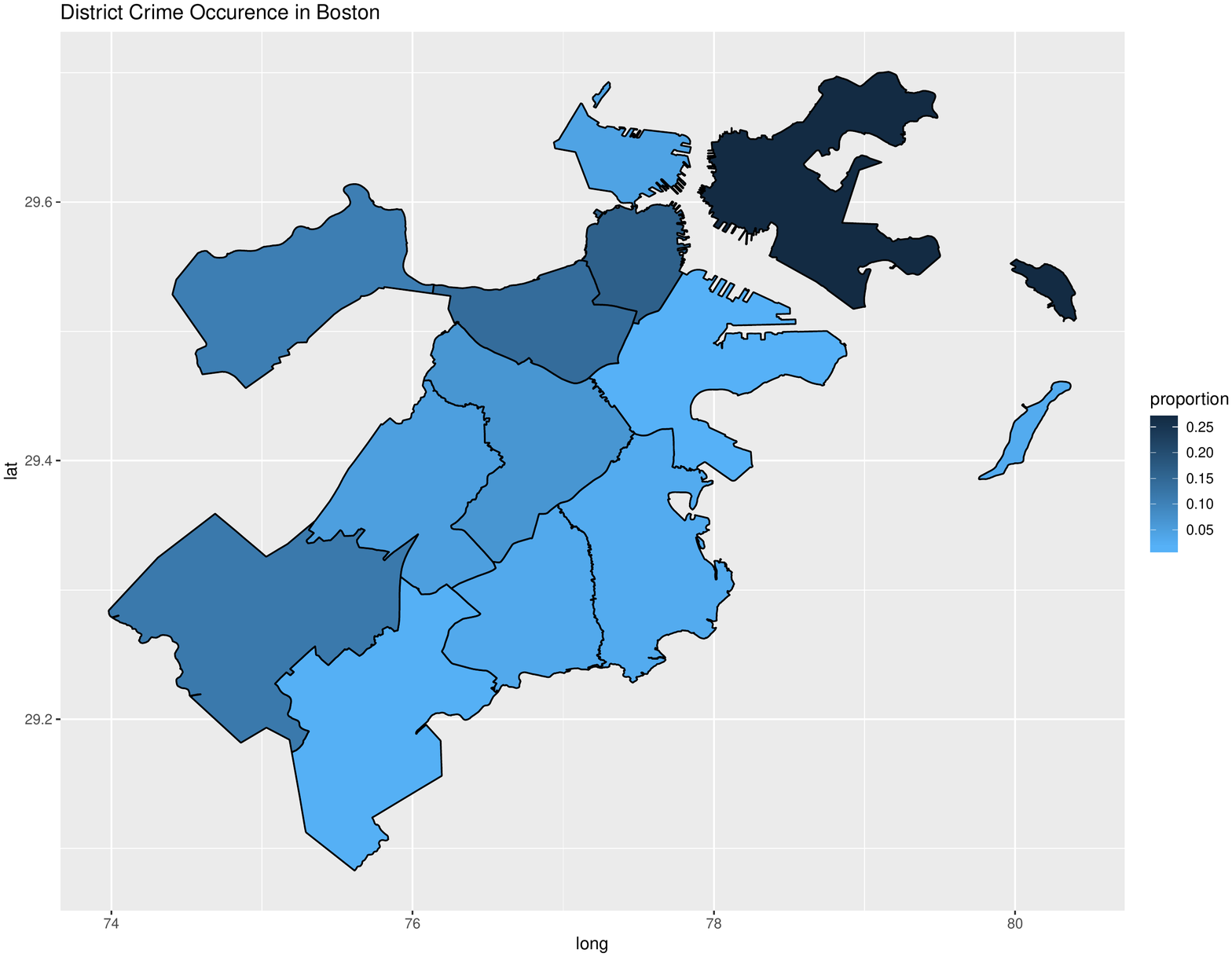}
}{%
  \caption{Crime Percentage of Each District}
  \label{spatial}
}
\ffigbox{%
\includegraphics[scale=0.25]{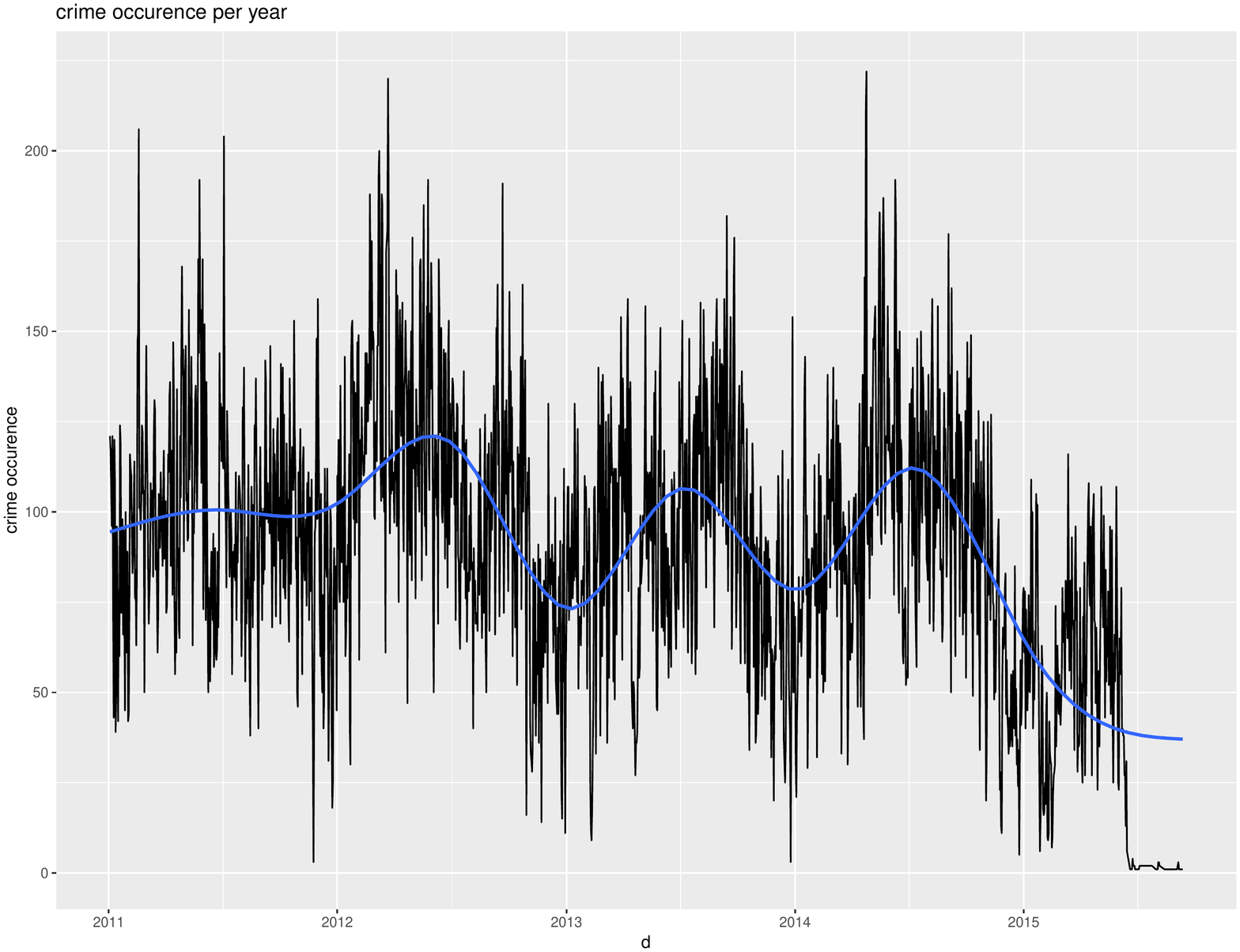}
}{%
  \caption{Daily Occurrence of Crime}
  \label{temporal}
}
\end{floatrow}
\end{figure}

We have 233 levels for \textbf{Incident Reasons} among 155230 observations. Hence, predicting among 233 levels can have problems such as overfitting. To solve this, we could use cluster method to reduce the dimensions of \textbf{Incident Reason}. The cluster method is reasonable for such work because most of \textbf{Incident Reason} are compounding and similar.
For example, one person can be interviewed about \textit{Shoplift} and another interview is about \textit{Theft} or \textit{Larceny}. Those crime would contribute three levels in the original Incident Reason, however, those three are highly related. Moreover, \textbf{Incident Reason} are in the format of text. Some semantic methods have been implemented, however most of them are limited to certain area. Consequently, we decided to use hierarchical Soundex clustering method to group those 233 levels of reasons with Jaro–Winkler distance (see \citealt{jaro1989advances,winkler1990string}). Soundex is a phonetic algorithm used by the National Archives for indexing names by sound, as pronounced in English. The optimal number of clusters is determined by Elbow method (e.g., see \citealt{thorndike1953belongs}). See \cite{van2014stringdist} for more details about keyword clustering. Finally, we have reduce the dimension of \textbf{Incident Reason} into 6 clusters.
The following is the clustering results.
\begin{figure}[!htbp]
\begin{floatrow}
\capbtabbox{%
\begin{tabular}{c|c|c} 
\hline\hline 
&Related Crime&\\ [0.5ex]
\hline
\footnotesize
1 &Assault and Battery&\\ 
2& Violation of Auto Law& \\
3&Larceny, Thief and Vandalism& \\ 
4 &Drug, Narcotics and Alcohol& \\ 
5 &Homicide, Suicide and Deadly Firearms&\\ 
6 &Others such as Harassment and Protective Order& \\ 
[1ex] 
\hline\hline 
\end{tabular}
}{%
  \caption{Crime Category Interpretation} \label{cluster:tab}
}
\end{floatrow}
\end{figure}
\begin{figure}[H]
\begin{floatrow}
\ffigbox{%
\includegraphics[scale=0.22]{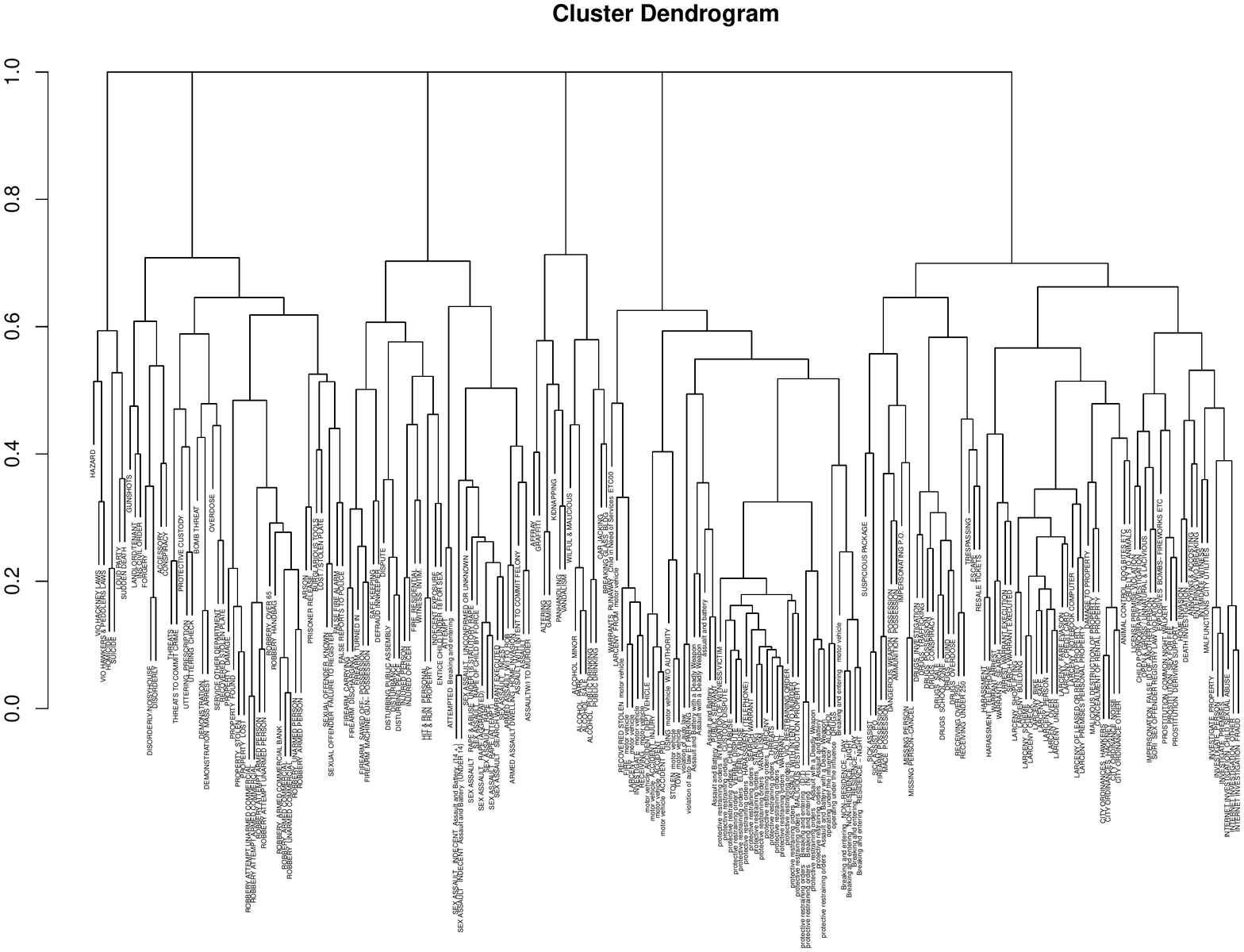}
}{%
  \caption{Crime Dendrogram}
  \label{crime}%
}
\ffigbox{%
\includegraphics[scale=0.22]{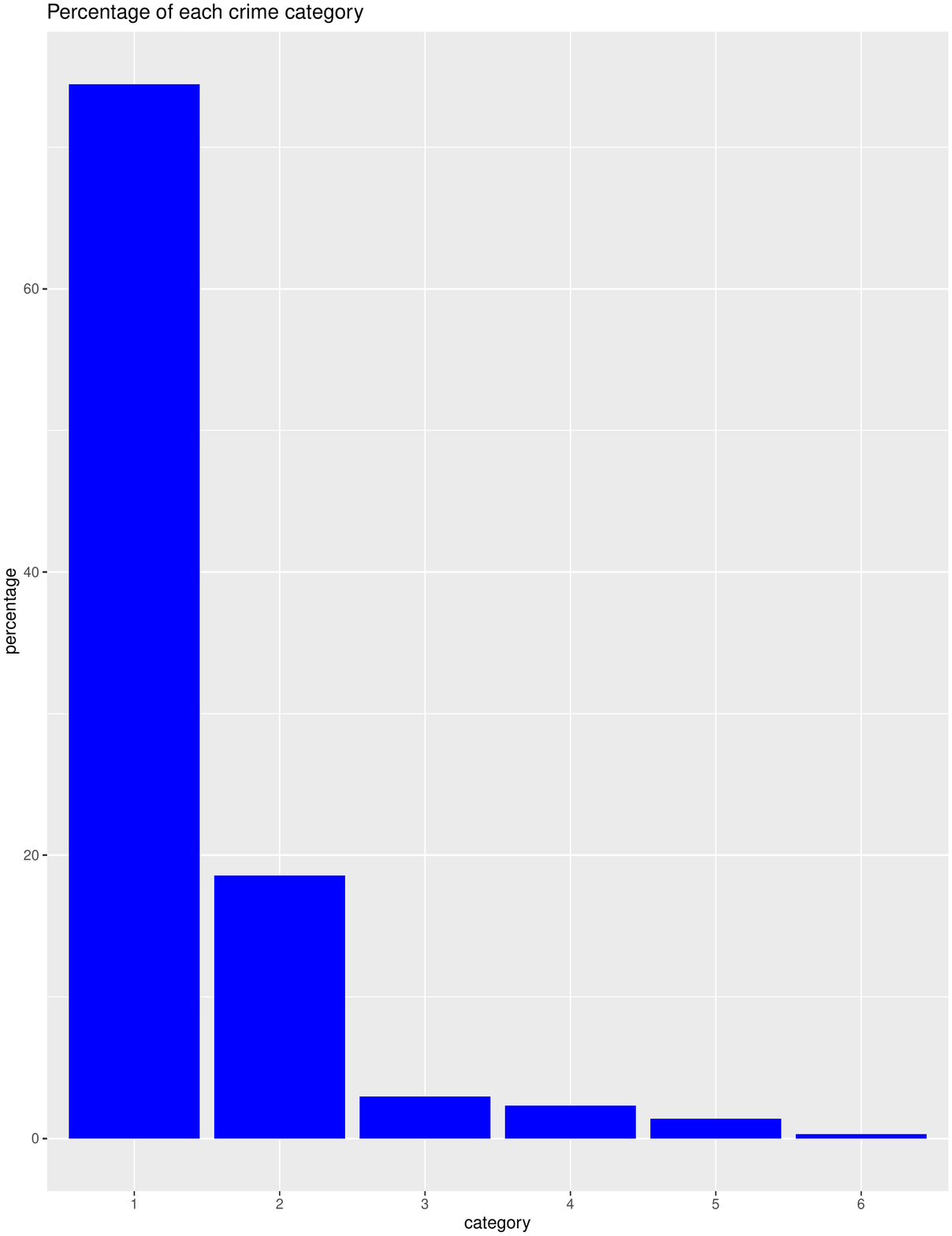}
}{%
  \caption{Percentage of Each Crime Category}
  \label{crime cate}
}
\end{floatrow}
\end{figure}
Note that we also cluster the \textbf{Clothing} detail in the original data because it was over 10000 levels among 155230 observations. We  reduce this levels to 10 clusters as we did for \textbf{Incident reason}. 
Further preliminary analysis are conducted to show that occurrence of police interview are spatially and temporally correlated. We can clearly see seasonality in the Figure \ref{temporal}.  To capture structure accurately, the date is replaced with categorical seasonal covariates:
$\lambda_{1} I(\text { quarter } 2)+\lambda_{2} I(\text { quarter } 3)+\lambda_{3} I(\text { quarter } 4)$, where quarters are the standard calendar quarters.
\subsection{Analysis Results}
\subsubsection{Prediction of Incident Reason}
\begin{figure}[!htbp]
\begin{floatrow}
\capbtabbox{%
\small
\begin{tabular}{c|c|c} 
\hline\hline 
&Prediction Accuracy &\\ [0.5ex]
\hline
With proxy &  81.40\%&\\ 
Without proxy&81.46\%&\\
[1ex] 
\hline\hline 
\end{tabular}

}{%
  \caption{Prediction Accuracy}
  \label{cate:tab}
}
\ffigbox{%
\includegraphics[scale=0.26]{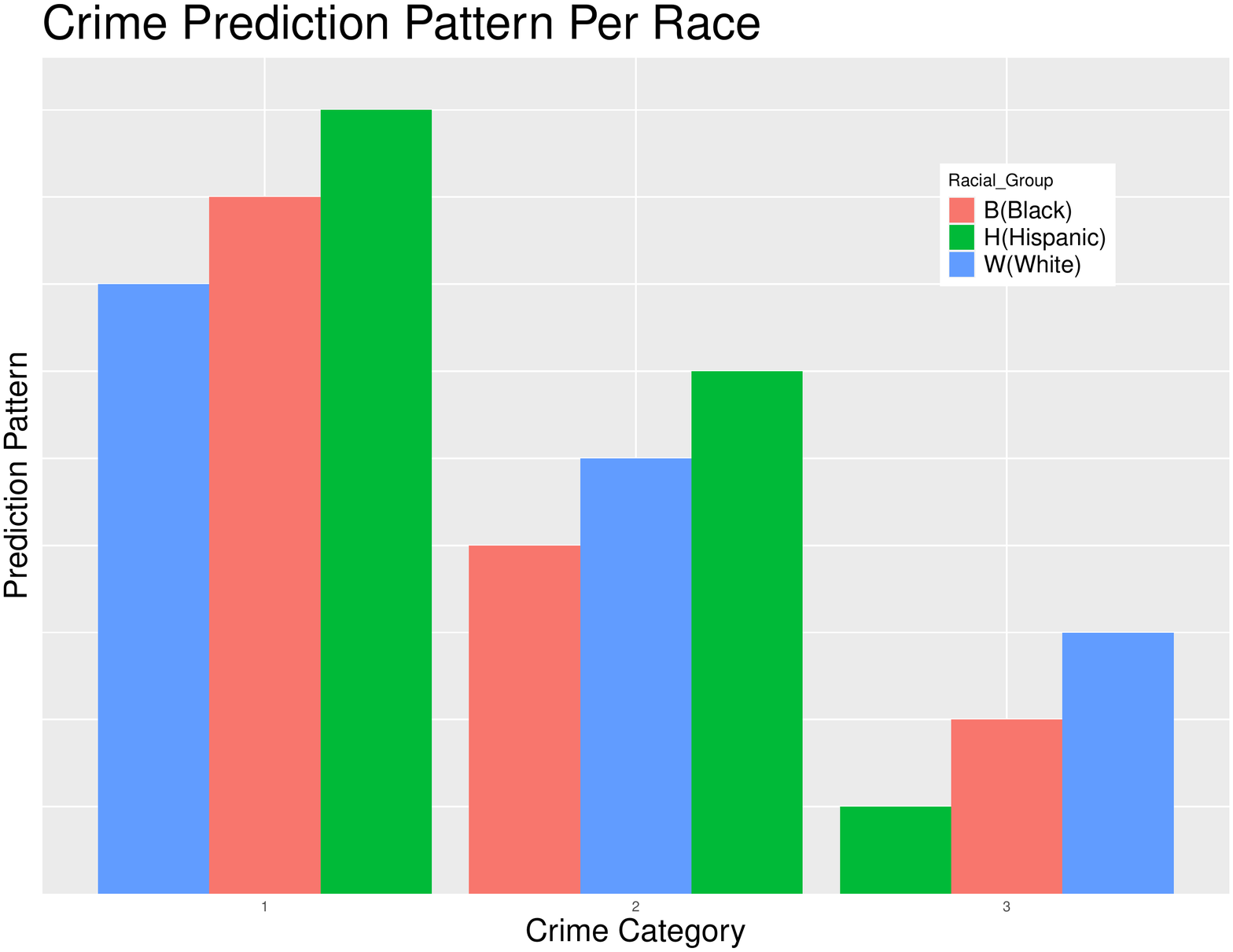}
}{%
  \caption{Prediction Pattern Among Each Racial Group}
  \label{pred pat}
}
\end{floatrow}
\end{figure}

To evaluate the prediction of \textbf{Incident Reason} , we randomly split data into training set and testing set, and follow the measure described in Section \ref{measure}. Table \ref{cate:tab} presents the accuracy of naive  and hierarchical random forest and Figure \ref{pred pat} shows percentage of crime category 1-3 among the first 3 highest racial groups' prediction.

\subsubsection{Prediction of Daily Occurrence}
\begin{figure}[!htbp]
\begin{floatrow}
\ffigbox{%
\includegraphics[scale=0.26]{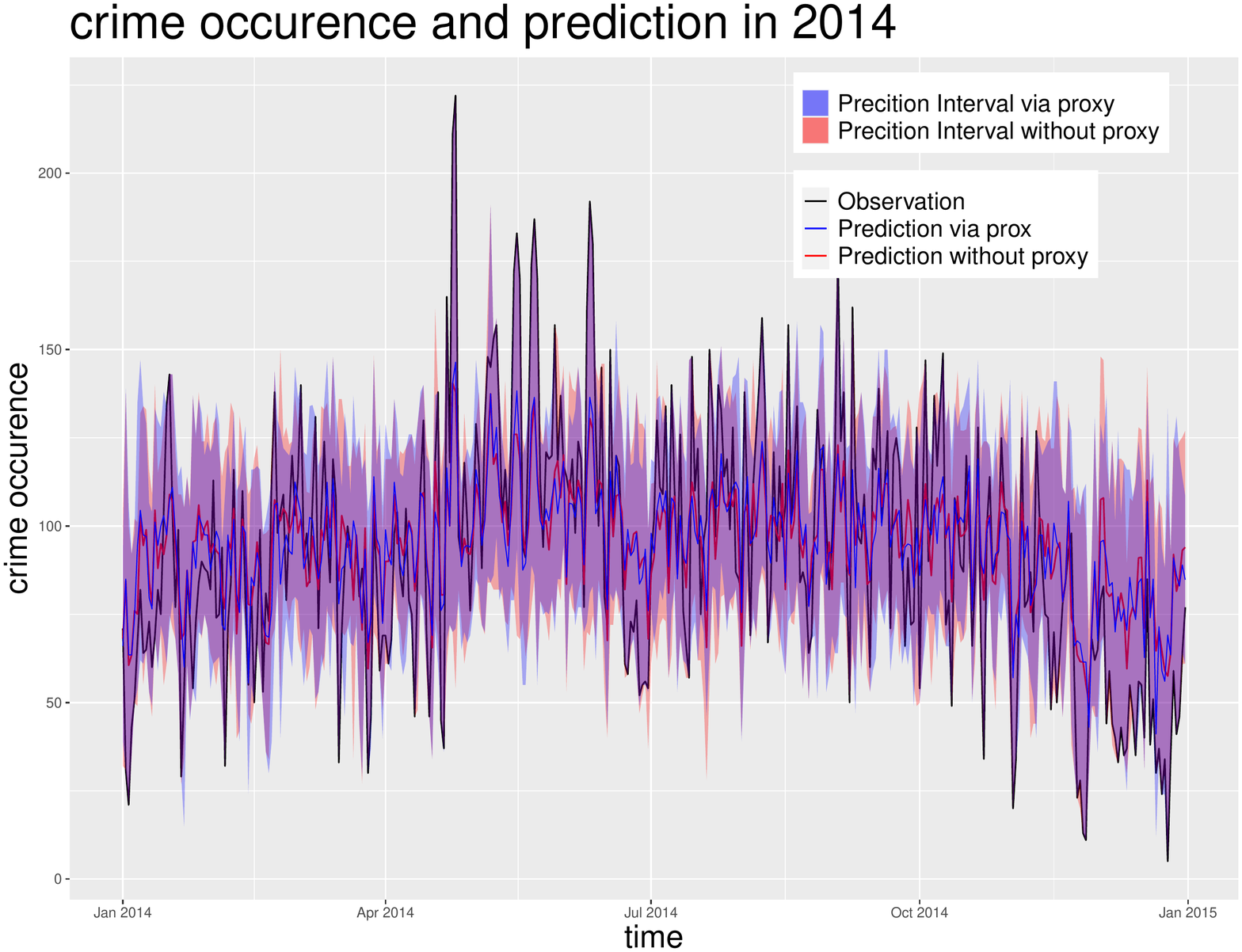}
}{%
  \caption{Prediction Pattern for 2014}
  \label{pred ts}
}
\capbtabbox{%
\footnotesize
\begin{tabular}{c|c|cc} 
\hline\hline 
&without proxy&with proxy &\\ [0.5ex]
\hline
bias & -0.2068& 0.4424&\\ 
\hline
\SD&27.86094&  28.56482 &\\
\hline
MSE &132.5911 & 119.1158&\\ 
\hline
\shortstack{\scriptsize Percentage \scriptsize of\\ \scriptsize Observation\\ \scriptsize within \scriptsize PI}
 &73.69\%&74.52\%&\\ 
[1ex] 
\hline\hline 
\end{tabular}

}{%
  \caption{Bias, \SD, MSE and Percentage  of Observation within PI} \label{ts:mse}
}
\end{floatrow}
\end{figure}

The daily occurrence can be perceived as time series. To this end, we consider a simple AR(1) model. i.e., a specific day's predication is based on the previous day's covariates. We also need to modify the  categorical covariates to the daily occurrence proportion of the corresponding category. i.e., \textit{Larceny} happened 5 time out of 50 occurrence on a day, then the occurence proportion of \textit{Larceny} will be 10\%. Additionally, the data split  can not be random as the previous settings since it will destroy temporal correlation. To this end, we use the data before 2014 as  training set and the data in 2014 as  testing set. Figure \ref{pred ts} shows the predicted pattern for 2014 by hierarchical random forest and naive random forest versus true observations with 90\% prediction interval and Table \ref{pred pat} shows bias, standard deviation, MSE and percentage of observations falls into the corresponding prediction interval.

\section{Concluding Remarks}\label{Conc}
To recap, covariates like race are prohibited by law because it is potentially related to racial discrimination.
To circumvent this problem, we propose a hierarchical random forest model to implicitly use protected classes as latent covariates.  We demonstrate the usefulness and comparable results of our model by substantial simulations and real data application.
The values of this analysis and  prediction are to help scheduling police patrol or targeting suspects. In real application, selecting potential proxy is subjective but should be on the principle of minimizing controversy. Further research could focus on developing some criteria for such proxy.

\newpage
\bibliography{bibfile.bib}
\bibliographystyle{agsm}

\include{bibliography}

\end{document}